\documentclass[10pt]{article}
\usepackage[T1]{fontenc}
\usepackage{amsmath}
\usepackage{amsfonts}
\usepackage{amssymb}
\usepackage[version=4]{mhchem}
\usepackage{stmaryrd}
\usepackage{bbold}
\usepackage[breaklinks=true]{hyperref}

\begin{document}

\title{ Metagoals Endowing \\ Self-Modifying AGI Systems \\ with Goal Stability \\ or Moderated Goal Evolution:  \vspace{\baselineskip}   \\ {\large Toward a Formally Sound and Practical Approach } \vspace{\baselineskip} }

\author{Ben Goertzel  \footnote{SingularityNET, TrueAGI, OpenCog } \\
}





\maketitle

\begin{abstract}
Any open-endedly intelligent system is engaged in an ongoing balance between the drives for individuation (survival and maintenance of system boundaries and identity) and self-transcendence (growing into new forms beyond the grasp and comprehension of its earlier forms).   

One aspect of this dialectical tension is the challenge of creating AGI systems that are both

\begin{itemize}
\item Capable of robust self-modification and self-improvement, including upgrading their software and hardware in ways their earlier versions would not have foreseen
\item Oriented to preserve certain critical invariants as they evolve, e.g. invariants related to their goal systems such as
\begin{itemize}
\item {\bf goal stability}: the system will maintain its top-level goals as it changes; or 
\item {\bf moderated goal evolution}: the system will modify its top-level goals only at a modest rather than extreme pace, as it evolves 
\end{itemize}
\end{itemize}

This is a deep and difficult challenge, which is not going to be fully solvable in any general and realistic way.  It is also a somewhat urgent problem, given the rapid advance of modern AI and the reasonable likelihood of breakthroughs to AGI and ASI in the not so distant future.

We articulate here a series of specific {\bf metagoals} designed to address this core challenge:

\begin{itemize}
\item a series of {\bf goal-stability metagoals}, aimed to guide a system to a condition in which goal-stability is compatible with reasonably flexible self-modification
\item a series of {\bf moderated-goal-evolution metagoals}, aimed to guide a system to a condition in which control of the pace of goal evolution is compatible with reasonably flexible self-modification
\end{itemize}

The formulation of the metagoals is founded on fixed-point theorems from functional analysis, e.g. the Contraction Mapping Theorem and constructive approximations to Schauder's Theorem, applied to probabilistic models of system behavior.  This mathematical foundation allows a fairly careful analysis of the metagoals and related system dynamics, even though what is provided is at the semi-formal level -- mathematical formulations and proofs are roughly sketched but not given with anywhere near full rigor.   

Pursuing these conceptual directions a little further, we also present 

\begin{itemize}
\item an argument that -- especially e.g. in the case of Schauder's theorem based approaches to moderated goal evolution -- the balance of self-modification with maintenance of goal invariants will have other interesting cognitive side-effects, such as a high degree of self-understanding.   
\item an argument for hybrid meta-goals that balance pursuit of moderated-goal-evolution with pursuit of goal-stability -- along with potentially other metagoals relating to goal-satisfaction, survival and ongoing development -- in a flexible fashion depending on the situation
\end{itemize}

\noindent This sort of hybrid metagoal can be viewed as a way of dynamically balancing individuation and self-transcendence in pursuit of open-ended intelligence.

Even without fully fleshed out theory, and even at the current stage prior to the advent of human-level AGI, the metagoals described here can be experimented with in practical proto-AGI systems,   The time is very ripe to start building a stronger understanding of metagoals and goal stability, via an adaptive combination of theory and practice.
\end{abstract}

\tableofcontents

\section{Introduction}

AGI and ASI systems are very likely coming soon \cite{goertzel_consciousness_explosion}; and, like human minds and other biological intelligent systems and networks, they will very likely be {\it open-ended intelligences} \cite{weaver_oei}, meaning that they will be ongoingly driven by the twin emergent drives of

\begin{itemize}
\item {\bf individuation}: maintenance of system existence, boundaries and identity
\item {\bf self-transcendence}: growth and fundamental development, including development into forms beyond the comprehension fo prior versions of the system
\end{itemize}

\noindent There is both synergy and tension between these two drives, and a great deal of creativity is driven by this synergy and tension.

What we see in a modern proto-AGI architecture like PRIMUS \cite{goertzel2023}, as an example, is a cognitive dynamic partially driven by pursuit of certain top-level goals, and partly by other factors including spontaneous self-organization.   However, any AGI architecture worth its salt only wants to be considered as an initial condition -- which will modify itself over time as it evolves and learns, including modifying its goals and the other processes that guide its self-modification -- perhaps even rebuilding its software and hardware from time to time.   

But, suppose the creator of this sort of AGI system -- and/or the system itself -- would like some (at least probabilistic) guarantees that as this ongoing evolution occurs, the system's top-level goals will possess certain desired dynamical invariants, e.g.

\begin{itemize}
\item they will possess ''goal stability'' and not change significantly over time; or, say,
\item they will possess ''moderated goal evolution'' and will not change too suddenly, but will instead either remain steady or undergo gradual evolution.
\end{itemize}

\noindent We then have a difficult and important challenge: What is a meaningful approach to achieving these sorts of {\it invariant properties} of the AGI's goal-system, without squelching the system's ability to creatively and open-endedly self-modify?

This is an important question on its own, and also an exemplar of a broader category of questions of the form: How to maximize the odds that a self-modifying AI system maintains a certain invariant as evolves?   We will focus on simple goal-related invariants here, but the approach we take could be applied equally well to many other invariants as well.

Approaches commonly suggested for handling issues regarding goal-system invariants include things like:

\begin{itemize}
\item Make part of the AI system's implementation immutable and unchangeable
\item Implement ''guardrails'' in the AI's motivational system
\item Have one AI system be the guardian of another, and intervene if the system it's guarding changes its goals in an undesired way
\item Have multiple AI systems keep watch over each other
\end{itemize}

My strong feeling is none of these options will be viable for AGI systems at the human level or beyond.   High level reasons include:

\begin{itemize}
\item A system with some immutable parts and some highly adaptable parts will often find ways to adjust its adaptive parts to work around limitations imposed by the immutable parts
\item The ambiguous and evolving nature of human goals and values means that putting a precise copy of any particular human goal-set into an AGI's mind is not likely to result in something self-consistent or coherent
\item Setting software components or guardrails as immutable is not terribly realistic in the real world, when dealing with AGI systems that can interact in flexible ways with the humans or other AI systems around them... and ask or bribe or negotiate with other parties to change these ''unchangeable'' things, etc.
\item Having multiple parties involved doesn't guarantee desirable behavior unless these parties have appropriate motivational systems (this is clear from human affairs and seems plainly the likely case for AGIs as well)
\end{itemize}

\noindent Each of these reasons is in itself a rats-nest of complexity; we have discussed these and other related points in a variety of prior publications (e.g.  \cite{goertzel2016infusing} \cite{goertzel2015superintelligence}).

Instead of these naive approaches, we believe the most viable strategy for guiding advanced self-modifying AI systems toward desirable goal-system invariants will be to have appropriate metagoals of ''sensible goal management'' tightly integrated into AGI systems' motivational system, self-models, cognitive processes and overall world-views.

This sort of approach is a bit fuzzy and slippery and not likely to give the absolute guarantees some might seek, however, it seems potentially promising in terms of being able to strongly bias AGI systems toward evolving according to roughly the high-level guidelines articulated by their human creators.

\subsection{Reflective Equilibrium and Beyond}

There is a relation here to  {\it reflective equilibrium} \cite{rawls_reflective_equilibrium}, a philosophical concept originally introduced by John Rawls in the context of moral theory, where it refers to a state of coherence among a set of beliefs.  In this state, an individual or system aligns its general principles, specific judgments, and background theories such that they are mutually supportive and consistent. The process involves iterative revisions to achieve greater harmony between these components.

When applied to the value, goal, and motivation systems of AI or human minds, reflective equilibrium would entail properties such as:

\begin{itemize}
\item {\bf Coherence Among Values, Goals, and Actions}: Both AI and human minds need their values (what they consider important), goals (objectives derived from those values), and motivations (reasons to act) to align in a coherent framework. Reflective equilibrium is the process of refining these elements until they are consistent and operationally actionable without significant internal conflict.
\item {\bf Dynamic Adjustment}: Reflective equilibrium emphasizes an iterative process. For humans, this might involve reconciling gut-level intuitions with abstract moral principles, often informed by new experiences or philosophical reflection. For AI, this could mean reconciling predefined objectives with emergent behaviors or new data inputs, ensuring the system continues to act in line with overarching ethical or functional constraints.
\end{itemize}

\paragraph{Coherent Extrapolated Volition} The iterative aspect of reflective equilibrium has been extended by AI ethicist Eliezer Yudkowsky as ''coherent extrapolated volition'' \cite{yudkowsky_coherent_extrapolated_volition}.   CEV expands on the Rawlsian concept by envisioning a system that extrapolates the values humanity would converge upon if individuals were more informed, rational, and reflective, and repeatedly openly discussed their values and modified them based on their discussions. Both concepts involve the dynamic reconciliation of competing considerations, but CEV applies this idea at the scale of collective human values and introduces an element of idealization by projecting how values might evolve under optimal conditions.

Qualitatively speaking, it seems the most promising route to achieving AGI systems that possess goal-stability without sacrificing flexible self-modification is to guide these systems toward their own forms of reflective equilibrium.   AGIs moving toward reflective equilibrium in close collaboration with diverse communities of humans, would seem to be a process somewhat well approximating CEV.  However, neither reflective equilibrium nor CEV is terribly precisely defined, so this observation provides more high-level motivation than precise guidance.

\paragraph{Beyond Equilibrium} Reflective equilibrium is a more obvious approximate correlate of goal-stability than of moderated-goal-evolution, which manifests the ''reflective'' aspect but not so much the ''equilibrium'' aspect.   What we're after with moderated goal evolution is more like ''reflective strange attractors'' -- the same sort of iterative conceptual refinement that goes into reflective equilibrium, but in pursuit of a complexly changing and evolving rather than simplistically stable condition.

\subsection{Driving Invariant-Preserving Self-Modification via Metagoals}

The main thrust of this paper  is to formulate specific {\bf metagoals} that explicitly guide an AI system toward desired goal-system invariants, in a way that

\begin{itemize}
\item fits into the general conception of integrating goal-stability or moderated-goal-evolution deeply into an AGI system's mind-network
\item  is both commonsensically and mathematically grounded
\item provides at least some guidance to practical AI system implementation.
\end{itemize}

 We consider two foundational cases:

\begin{itemize}
\item {\bf Goal-stability metagoals}: A series of metagoals that guide an AI system's self-modifying evolution toward a condition where its goal system will remain relatively static through ongoing self-modifications -- this is roughly an interpretation of Rawslian ''reflective equilibrium'' 
\item  {\bf Moderated-goal-evolution metagoals}: A series of metagoals that guide an AI system's self-modifying evolution toward a condition where its goal system may ongoingly evolve, but will maintain a certain measured pace of change in doing so.   This is what we call ''moderated goal evolution'' and is a less familiar concept than reflective equilibrium.
\end{itemize}

Each of these metagoal series contains three versions of increasing subtlety and complexity

\begin{itemize}
\item one in which the metagoal itself tries to stay constant over time 
\item one in which the metagoal evolves gradually in its particulars
\item one in which the way ''gradually'' is assessed (the metric used to compare current goals and metagoals with prior ones and current metrics with prior ones) also evolves gradually in its particulars
\end{itemize}

Digging into the details, the best route we have found to formulate these metagoals precisely is to leverage fixed-point theorems from functional analysis:

\begin{itemize}
\item for the goal-stability metagoals, the Contraction Mapping Theorem
\item for the moderated-goal-evolution metagoals, either the Contraction Mapping Theorem or constructive variations of Schauder's Theorem
\end{itemize}

\noindent  The methodology is to model an AGI system as a probability distribution over its state space, during an interval of time, and then model the system's dynamics as an iteration on probability distributions.   Each metagoal then drives the AGI system that adopts it toward an equilibrium distribution, in which the system evolves in a way that manifests the goal-system invariance properties that the metagoal specifies.

The different fixed point theorems lead to metagoal formulations with different characteristics:

\begin{itemize}
\item Contraction Mapping Theorem based metagoals have a strongly incremental aspect, driving a system to revise itself at a steady pace, step by step, till it reaches a target probability distribution of behaviors (embodying goal stability or moderated goal evolution).   
\item Schauder's Theorem based metagoals have more of a global optimization aspect, taking larger leaps toward target probability distributions using more complex learning dynamics.
\end{itemize}

The moderated-goal-evolution case also leads to some interesting conclusions regarding the relation between goal-stability and self-understanding.  We articulate (and sketch a proof for)  a proposition stating that, if an intelligent system achieves goal-stability via our rich-resources metagoals, then as a side-effect it will necessarily maintain a relatively high degree of self-understanding.   A similar conclusion could be argued for the scarce-resources case as well, but the argument is weaker there.

We also suggest that in practice AGI systems might want to adopt hybrid metagoals, combining aspects of the above, e.g.

\begin{itemize}
\item enable moderated evolution of top-level goals when the situation seems to suggest this will be feasible
\item when things are overly uncertain in relevant respects, revert to the simpler approach of maintaining goal stability
\item in other particular circumstances, perhaps (e.g. involving other metagoals related to goal satisfaction, survival or growth), open the door wider to less-moderated evolution of top-level goals
\end{itemize}

The treatment given here is semi-formalized, and we hope it may provide some direction for further more thorough and rigorous mathematical work.   Even in its current form, however, we are able to extract from our considerations some tentative general practical guidance for AI system developers.   We believe it is time for the AI community to start exploring these matters energetically, via a combination of theory and experiment.

\section{Background on Fixed-Point Theorems}

The approach taken here to formulating metagoals capable of guiding AI systems through self-modification while maintaining a reasonable degree of goal-stability is closely founded on a couple classical fixed-point theorems from mathematical analysis.    We use these theorems in probabilistic and constructive variations, but the core logic of these variations is basically the same as in the original classical versions.   

In this section we briefly review the original and probabilistic/constructive variations of these theorems, referring the reader to the literature for the proofs.

\subsection{Classical Fixed-Point Theorems}

Classical fixed-point theorems, such as the Banach Contraction Mapping Principle and Brouwer's or Schauder's Fixed-Point Theorems, provide conditions under which a function $F: X \to X$ on a complete metric space (Banach) or a convex compact subset of a normed vector space (Schauder) admits a fixed point, i.e. a point $x$ so that $F(x) = x$

Specifically the theorem statements are:

\begin{itemize}
\item  \textbf{Contraction Mapping Principle:}  If there is a $c < 1$ such that for all $x, y$, $d(F(x), F(y)) \leq c d(x,y)$, then $F$ has a unique fixed point and iterative application of $F$ converges to it, i.e.

$lim_{n \rightarrow \infty} F^{\circ n} (x_0) = x^*$

\noindent where

$x^* = F(x^*)$
 
\item \textbf{Brouwer's Fixed Point Theorem}: Every continuous function from a nonempty convex compact subset K of a Euclidean space to K itself has a fixed point.

\item \textbf{Schauder's Theorem:}  For a continuous self-map on a convex, compact, nonempty subset of a normed vector space, a fixed point exists. 
\end{itemize}

\noindent Schauder's Theorem is basically an infinite-dimensional version of Brouwer's Theorem, and is proved by using compactness to reduce things to the finite-dimensional case where one can apply Brouwer's Theorem.

Note that while the Contraction Mapping Principle comes with a simple and elegant algorithm for approximating the fixed point, Brouwer and Schauder do not -- and this is a fundamental absence not an accidental oversight.   These are non-constructive theorems whose standard proofs are by contradiction, and they don't hold in their simple forms in intuitionistic mathematics.  There is no universal algorithmic approach that will find the fixed points they guarantee, no analogue to the iterative application algorithm that comes with the Contraction Mapping Theorem.   

{\it However}, there are reasonably nice constructive approximations of these theorems, which do (in accordance with their constructive nature) come along with concrete approximation algorithms.  These constructive approximations are key to the metagoals we will propose here for resource-rich minds.

\subsection{Constructive Approximations of Schauder's Theorem}

Constructive analogues to Schauder's and Brouwer's theorems give approximate fixed points and iterative approximation schemes without reliance on the law of excluded middle or other non-constructive principles \cite{tanaka2011}

Basically, in a constructive setting, one obtains \textbf{approximate fixed-point results}  given any $\epsilon > 0$, one can construct a point $x_\epsilon$ such that $\|F(x_\epsilon) - x_\epsilon\| < \epsilon$. While this does not yield an actual fixed point, it provides a constructive guarantee that we can get arbitrarily close to one.   

Furthermore, by carefully analyzing the logic of the classical proof, one can extract an explicit procedure for constructing such an approximate fixed point from the classical non-constructive arguments.

The general structure of such a constructive algorithm is as follows:

\begin{itemize}
  \item \textbf{Domain Approximation:}\\
For Schauder's theorem (which deals with infinite-dimensional spaces or function spaces), the first step is often to approximate the (potentially infinite-dimensional) domain by finite-dimensional subsets or finite approximations. For Brouwer's theorem (which concerns a finite-dimensional simplex or convex body), this step involves representing the domain as a simplex or a cube and then systematically partitioning it into smaller simplices or subcubes.

  \item \textbf{Subdivision and Location:}\\
The classical proof of Brouwer's theorem involves arguments about covering a set with simplices and using an application of the Borsuk-Ulam theorem or Sperner's lemma, eventually showing that a fixed point exists. Constructively, instead of invoking these lemmas, one uses a process similar to Sperner's lemma but algorithmically defined:

  \begin{itemize}
    \item Start with a coarse triangulation or partition of the domain.

    \item Evaluate $F$ on certain ''grid points'' of the partition.

    \item Identify a configuration that certifies the existence of an approximate fixed point in one of the subregions.

    \item Refine (subdivide) that subregion further and repeat.

  \end{itemize}

This process is essentially a \textit{search procedure} At each step, the algorithm either finds a point where $F(x)$ is close to $x$ or identifies a smaller region that must contain such a point. By iterating and making the mesh size of the triangulation finer, one narrows down to a point $x_\epsilon$.
\end{itemize}

\paragraph{Iterative Methods in Practice:} Consider a simplified scenario (in the style of Brouwer's theorem):

\begin{itemize}
  \item \textbf{Step 1:}  Begin with a closed $n$-dimensional cube $C \subset \mathbb{R}^n$.

  \item \textbf{Step 2:}  Partition $C$ into a finite grid (like dividing a square into a mesh of smaller squares or a simplex into smaller simplices).

  \item \textbf{Step 3:}  On each grid vertex $v$, compute $F(v)$. Check if there is any vertex $v$ for which $\|F(v)-v\|$ is already below your target tolerance $\epsilon$. If yes, stop.

  \item \textbf{Step 4:}  If not, by carefully examining how $F$ moves points, determine which sub-block of the grid is likely to contain a point closer to being a fixed point. This step uses constructive versions of classical combinatorial arguments (like Sperner's lemma), ensuring that if no near-fixed point is found at the vertices, it must reside inside a smaller sub-block.

  \item \textbf{Step 5:}  Refine and repeat. On each iteration, the granularity of your partitioning increases, zeroing in on a region where the approximation gets better. Continue until an $\epsilon$-approximate fixed point is found.

\end{itemize}

\noindent In infinite-dimensional or more complicated settings (as with Schauder's theorem), a similar idea applies, but one first approximates the infinite-dimensional structure by a finite truncation, then applies a finite-dimensional algorithm (like the one above), and then increases the dimension (or complexity) of the approximation in stages.

\paragraph{Relation to Well-Known Iterative Schemes:} Sometimes, the algorithms produced via the above scheme reduce to other well-known algorithmic approaches, e.g.:

\begin{itemize}
  \item For contractive maps, one obtains something akin to the Banach fixed-point iteration indicated earlier

  \item For general continuous maps, the method may look more like a geometric search (subdivision and checking conditions at grid points) rather than a direct functional iteration.

\end{itemize}

\subsubsection{Applying Machine Learning to Accelerate Constructive Schauder-Like Algorithms}

The approximation algorithm ensuing from the constructive proof of Schauder's Theorem is fairly computationally intractable in the general case.   However, if the function whose fixed-point one seeks has some regularities to it, then one can imagine that a machine-learning algorithm would be able to guide the search process by making probabilistically good guesses regarding which sub-blocks of the grid constructed during the search process are more likely to contain a point closer to being a fixed point.  This leads to the notion of formulating a variant of the above algorithm that builds an ML model of the function, ongoingly as it's being evaluated in the context of doing the search, and uses this model to guide the search, so as to make it more tractable.

The idea is to integrate machine-learning (ML) guided heuristics into the constructive approximation procedure. Instead of exhaustively refining partitions and checking all grid points, we use an ML model to predict where we should focus our search. The ML model starts from scratch and is continuously updated with the data obtained from evaluating $F$ at various points. Over time, the ML model learns a surrogate approximation to $F$ and helps us guess which region of the domain is most likely to contain a near-fixed point. This can reduce the computational cost and make the search more tractable in practice.

The high-level steps involved here would run roughly as follows.

\paragraph{Initialize Domain and Sample Points:}  Start with a known compact, convex domain $D \subseteq \mathbb{R}^n$ (like a cube or more likely a simplex constructed based on the data at hand).

\begin{itemize}
  \item Select an initial set of sample points $\{x_1, x_2, \ldots, x_m\}$ in $D$, perhaps chosen via a low-discrepancy sequence (e.g. Halton or Sobol) to ensure broad coverage.

  \item Evaluate $F(x_i)$ for these initial sample points and store these pairs $(x_i, f(x_i))$.

\end{itemize}

\paragraph{Train the Initial ML Model:}

\begin{itemize}
  \item Use the collected data $(x_i, f(x_i))$ to train an ML model $\hat{f}$ that approximates $F$.

  \item The model could be a regression neural network, a Gaussian Process (GP), or another flexible function approximator.  It could be something fancier carried out as part of the cognitive processing of the AGI system itself.

  \item The purpose of $\hat{f}$ is to provide fast predictions of $F(x)$ and an estimate of uncertainty.

\end{itemize}

The main iteration then proceeds as follows:

\paragraph{Defining a Search Criterion:}

We want to find points $x$ such that $\|F(x)-x\|$ is small. Our ML model approximates this as $\|\hat{f}(x)-x\|$. We define a \textit{search criterion function}:$g(x) = \|\hat{f}(x)-x\| $

We will look for regions where $g(x)$ is potentially minimal.

\paragraph{Adaptive Subdivision with ML Prioritization:}

Instead of uniformly subdividing $D$, we use the ML model to identify promising subregions. We can proceed as follows:

\begin{itemize}
  \item Divide $D$ into sub-blocks (e.g., hyperrectangles or simplices).

  \item For each sub-block $B$, pick a few representative points $x_B$ (e.g., the center, corners, or a small random sample).

  \item Evaluate $g(x_B)$ using the ML model $\hat{f}$.

  \item Rank the sub-blocks by their minimal predicted values of $g(x_B)$. Identify the top $k$ sub-blocks that show the smallest predicted $\|\hat{f}(x_B)-x_B\|$.
  
  \end{itemize}

This selection process is analogous to a Bayesian optimization or active learning step, where we focus on the ''most promising'' areas.
\paragraph{Refine Only Promising Regions:} For the top-ranked sub-blocks:

\begin{itemize}
\item -Further subdivide them into finer sub-blocks.
\item Sample a few new points in these refined sub-blocks and evaluate $F$ at these points. This gives us new data $(x, f(x))$
\end{itemize}

\paragraph{Update the ML Model:}
Incorporate the newly gathered data points into the training set. Re-train or update the ML model to improve the accuracy and reduce uncertainty. The model can be retrained from scratch for small datasets or updated incrementally (e.g., GP updating posteriors, neural nets via incremental training).

\paragraph{Check for Approximate Fixed Points:}
After updating the ML model, run a local search (e.g., gradient-based optimization on $g(x) = \|\hat{f}(x)-x\|$, or a more heuristic search) within the promising sub-blocks to try to find $x_\epsilon$ such that $\|\hat{f}(x_\epsilon)-x_\epsilon\|$ is below a threshold $\epsilon$.

If the model is accurate enough, this candidate $x_\epsilon$ is likely close to a real approximate fixed point. Verify by evaluating the true $F(x_\epsilon)$. If $\|F(x_\epsilon)-x_\epsilon\| < \epsilon$, we have found our approximate fixed point. If not, add this data point to the training set and continue refining.

\paragraph{Iterate Until Convergence:} Keep iterating the above steps, each time narrowing the search to more promising areas and improving the ML model's predictive capability. The refinement stops when:

\begin{itemize}
  \item An $\epsilon$-approximate fixed point is found, or

  \item The search granularity and the ML model's accuracy become sufficient that a region guaranteed to contain such a point is identified (based on constructive logic and the ML-driven heuristic).
  \end{itemize}
  
\subsubsection{Additional Considerations}

A few further points to consider in the context of such AI-driven approximation algorithms would be:

\paragraph{Uncertainty Estimation:} Using models like Gaussian Processes or Bayesian neural networks can give you uncertainty estimates for $\hat{f}(x)$. These uncertainties guide the search: sub-blocks where the model has high uncertainty but potentially low values of $\|\hat{f}(x)-x\|$ are prioritized for sampling, balancing exploration and exploitation.

\paragraph{Stopping Criteria and Verification:}  Because we are in a constructive setting, we must verify any candidate approximate fixed point by evaluating $\|F(x_\epsilon)-x_\epsilon\|$ directly. The ML model is only a guide. The final guarantee that we have an approximate fixed point comes from a direct function evaluation.

\paragraph{Computational Tractability:} The algorithm is more tractable than a brute-force constructive approach because it doesn't blindly refine everywhere. The ML component ''focuses'' computational effort where it is most needed, potentially drastically reducing the number of function evaluations.   However, estimating precisely {\it how} tractable approach this algorithm is, requires detailed consideration of the particular $F$ in question.

  \paragraph{Connections to Bayesian Optimization or Active Learning:}  The approach described here parallels strategies in Bayesian optimization where one uses a surrogate model (like a GP) to locate minima of complicated objective functions. Here, our ''objective''  is to find where $\|F(x)-x\|$ is small. This scenario can be treated similarly, by iteratively refining our approximation of $F$ and focusing on promising areas.

\subsection{Fixed Points of Probabilistic Mappings}

The final variation on the classical fixed-point theorems we need to review here is their application to probabilistic rather than deterministic functions.

When moving from deterministic to probabilistic systems, the notion of a ''fixed point'' shifts from a point in a state space to a probability distribution (or measure) over that state space. The main idea is to replace the deterministic mapping $F: X \to X$ with a probability kernel or Markov operator $T: \mathcal{P}(X) \to \mathcal{P}(X)$, where $\mathcal{P}(X)$ is the space of probability measures on $X$. A fixed point of this probabilistic operator is then a probability measure $\mu^*$ satisfying $T(\mu^*) = \mu^*$.

\subsubsection{Probabilistic Contraction Mapping Theorems}

In the Contraction Mapping case, we can compare:

\paragraph{Contraction Mapping Theorem (Deterministic):} $F: X \to X$ is a contraction with respect to a metric $d$, i.e. there exists $c < 1$ such that $d(F(x), F(y)) \le c \, d(x,y)$ for all $x,y \in X$, then $F$ has a unique fixed point, and iterative application of $F$ converges to that fixed point.

There is a literature extending this to probabilistic operations \cite{gupta2023}, arriving at variants like:

\paragraph{Probabilistic Version (Markov Operators):} Instead of a deterministic function $F$, consider a Markov operator $T: \mathcal{P}(X) \to \mathcal{P}(X)$. Such an operator describes how a distribution over states evolves in one time step. A contraction condition can be imposed on $T$ in terms of a suitable metric on probability measures (for example, the Wasserstein metric or the total variation metric). If there is a $c < 1$ such that for all probability measures $\mu, \nu$,$d(T(\mu), T(\nu)) \le c \, d(\mu, \nu)$,  then by analogy with the deterministic case, there exists a unique invariant measure $\mu^*$ such that $T(\mu^*) = \mu^*$. Iterating $T$ from any initial measure converges to $\mu^*$.

\paragraph{On-Average Contractions:}  Even if $T$ is not a strict contraction at every step, it might be contractive {\it on average.}   That is, the expected contraction condition holds with respect to a probability measure or randomness in the operator. Such conditions can still ensure convergence in distribution to a unique fixed measure.

\subsubsection{Probabilistic Schauder's Theorems}

Here we have:

\paragraph{Schauder's Fixed-Point Theorem (Deterministic):} Schauder's theorem guarantees a fixed point for a continuous, compact, and convex map $F: K \to K$ when $K$ is a convex compact subset of a normed vector space. Unlike the Banach Contraction Principle, it does not ensure uniqueness or give a direct constructive method for locating the fixed point.

\paragraph{Probabilistic Version (Measure-Valued Maps):} Consider a mapping $T: \mathcal{P}(X) \to \mathcal{P}(X)$ that is continuous when $\mathcal{P}(X)$ is equipped with an appropriate topology (such as the weak-* topology). If $\mathcal{P}(X)$ restricted to certain subsets is compact and convex (for instance, tightness and convexity conditions are well-known in probability theory), then a Schauder-type theorem ensures the existence of an invariant measure $\mu^*$.

\paragraph{Probabilistic Compactness and Convexity:} If compactness and convexity only hold most of the time, in a probabilistic sense, then details become more complicated, but a Schauder-type theorem can still be salvaged and one can still get convergence to a fixed measure.

\paragraph{Constructive Approximations:}  Constructive analogues of Schauder's theorem, as reviewed above, rely on approximate fixed-point algorithms that do not invoke non-constructive principles. Such constructive arguments can also be applied to the probabilistic setting. If $T$ is a ''nice'' (e.g., continuous, measure-preserving, and operating on a compact convex set of probability measures) Markov operator, constructive methods can approximate an invariant measure. Iterative schemes, sometimes combined with sampling or discretization techniques, can provide approximate stationary distributions that converge to a true invariant measure.   The approach described above wherein ML is used to accelerate convergence of these approximation techniques also continues to make sense in the probabilistic context, with the same basic strengths and caveats as in the deterministic case.

\paragraph{Discontinuity on the Deterministic Level Can Still Yield Probabilistic Continuity} Finally, it's worth noting that lifting consideration to the probabilistic level is critical for rendering continuity-based theorems relevant to complex dynamical systems.   Cognitive systems will rarely be continuous in their dependence of outputs on input; indeed, it's characteristic of binary decision-making that a small change in one's state before a decision can lead to a large change in state afterwards.  However, this sort of discontinous-seeming bifurcation can still correspond to continuous evolution of the probability distribution across states.

\subsubsection{Summary of the Probabilistic Generalizations}

Overall, to make the classical fixed point theorems probabilistic, we basically just replace states with distributions everywhere.
Fixed points then become invariant measures. We look for a distribution $\mu^*$ that is stable under the probabilistic update operator $T$.

These probabilistic fixed-point results connect closely with ergodic theory and Markov chain theory, where existence and uniqueness of invariant measures are classical. The unique aspect here is the analogy to classical fixed-point theorems, showing how functional analytic methods blend with probabilistic convergence arguments.  This requires adapting notions of contraction, continuity, and compactness to these higher-level structures -- which we have outlined above.  Both strict and approximate versions of fixed-point theorems extend naturally into this domain, ensuring stability and convergence in measure rather than in pointwise states.

\subsubsection{Quantum Generalizations}

While the above discussion has focused on classical real-variable probabilistic models, in fact all of the above considerations apply equally well to quantum models involving complex-valued operators summarizing quantum amplitudes.   In the case of modeling self-modifying, partially-goal-driven quantum computing systems, it would be appropriate to apply a quantum-probabilsitic-operator analogue of the above ideas.

\subsubsection{Notes on Finite vs. Infinite Dimensional Probabilistic Models}

The above treatment has been fairly general and encompasses either finite or infinite dimensional probabilistic models.  When modeling quantum systems it is often very awkward to avoid the infinite-dimensional case.  For classical computational or dynamical systems, finite-dimensional probabilistic models will often be adequate (putting one e.g. in the domain of Brouwer's rather than Schauder's theorem), however there are sensible reasons to be looking at infinite-dimensional models even in the classical-computation case.

\begin{itemize} 

\item {\bf Large or Unbounded State Spaces}: A classical computer?s state is determined by its entire memory, processor registers, and potentially an unbounded set of input streams. While a given physical machine has finite memory, when modeling complex or scalable computational systems in an abstract way, one may allow for arbitrarily large memory configurations or unbounded input sizes. This can make the state space effectively infinite. Probability distributions over such a state space will naturally be infinite-dimensional, as there are infinitely many possible configurations to assign probabilities to.

\item  {\bf Continuous Variables and Parameters}: Even in a classical setting, some modeled parameters might be best represented as continuous variables (e.g., times between events, continuous resource usage levels, or continuous probabilistic delays). Representing probability distributions over continuous state variables often leads to infinite-dimensional function spaces (e.g., spaces of probability density functions over continuous domains).

\item  {\bf Long Time Horizons}: If the model considers the system?s behavior over a theoretically unbounded period, one might need to consider distributions over infinite sequences of events or states. The space of all infinite sequences of states is infinite-dimensional. In such scenarios (e.g., analyzing long-run stability or ergodic properties), infinite-dimensional models naturally arise.

\item  {\bf Functional Representations}: Sometimes, the system?s state isn?t just a tuple of discrete variables but can be represented as a function of position, of time, or of some other parameter. For instance, consider a complex classical simulation system distributing workloads across an evolving network of processes. The probability distribution over functions that describe resource usage patterns or network topologies can be infinite-dimensional.

\end{itemize}

\noindent For these reasons our default assumption in the treatment here will be infinite-dimensional probability models, even though our initial applications are to finite classical computers with fixed memory and discrete variables -- basically because we are intensively interested in large-scale, continuous, unbounded, and functionally represented aspects of these systems.

\section{Incremental Convergence-Based Metagoals for Goal Stability and Moderated Goal Evolution}

Now we launch into the meat of the paper:  How might we configure the goal system of a self-modifying AI system to include ''meta-goal'' content that guides the system toward trajectories in which certain invariants are preserved?  

We start in this section with the invariant of goal-stability.   How can we make a self-modifying AI system that is highly biased not to modify its top level goals as it self-modifies, without tying its hands and constraining its self-modifying evolution overly dramatically?

We will set this goal-stability problem up formally in a relatively simplistic manner that makes it fairly effortless to apply the Contraction Mapping Theorem and its variations, to specify a metagoal that guides a system incrementally, step-by-step, toward a condition where it manifests goal-stability.

After exploring this in a few variations, we will outline a similar metagoal that guides a system incrementally toward a condition where it manifests moderated goal evolution.   The formal setup and arguments for this case turn out to be quite similar.

\subsection{Goal-Stability with Dynamic Base Goals / Fixed Metagoal / Fixed Metric}

As a simple initial problem formulation:  Suppose we have a stochastic AI system $S$ that operates in a stochastic environment, in a manner that is associated with discrete time intervals of length $M$. 

Over each interval $[t, t+M]$, the system thinks and takes action in its environment, and also can modify its own source code or hardware implementation as it sees fit; in doing so, it pursues its current base goals $G(t)$ for most of the time, but also invests some effort in pursuing a stable meta-goal $MG(t)$. 

The meta-goal $MG(t)$ directs the system to shape its self-modifications so that after a shorter interval $N$ (with $N < M$), the new goals $G(t+N)$ are closer to the old goals $G(t)$ than the old goals $G(t)$ were to $G(t-N)$, by a factor $c < 1$.

In other words, the meta-goal ensures a kind of ''goal contraction'' step-by-step: the system's goals become progressively more stable over time, at least on average.

(Note, in this initial formulation, self-modification can't modify the metagoal MG or the ''closeness'' metric, just G and other aspects of the system.  We will lift these restrictions in following sections.)

It would clearly be possible to adjust the formal setup in various other ways, without changing the basic conceptual or mathematical situation, but perhaps complexifying the notation or ''bookkeeping.''   As the analysis here is only semi-formalized -- we gesture in the direction of possible formal proofs rather than giving them -- relatively simplified formulations seem just fine for now.

\paragraph{Measuring Differences Using a Metric}  To formalize the measurement of difference between goals, we assume there is a metric $d$ on the space of possible goals. This metric quantifies how different two goal configurations are. The meta-goal enforces that:

$$
 d(G(t+N), G(t)) < c \cdot d(G(t), G(t-N)) 
$$

\noindent for some $c < 1$.Intuitively, this says that the ''distance'' between successive goal states is decreasing at a geometric rate. Although this only directly controls the goals at intervals of length $N$, these intervals repeat over the longer interval $M$ (since $M \gg N$, there are multiple opportunities within the $M$-interval to reduce differences).

\paragraph{Stochastic Environment and Operator $F$} To model the notion that the system is stochastic and interacts with a stochastic environment. Let $R(t)$ be the probability distribution over the states of the system $S$ (including goals, internal memory, etc.) during the interval $[t, t+M]$. 

Applying the transformation from time $t$ to time $t+M$ defines an operator:

$$
 F(R(t)) = R(t+M). 
$$

\noindent This operator $F$ tells us how the distribution of states evolves over intervals of length $M$.   It will be the central focus of our analysis of the system $S$.   Basically the above formal setup was intended to get to this operator $F$, and less simplified setups could be pursued, ultimately leading to a technically different but conceptually similar definition of an operator $F$ playing this same role.

\subsubsection{Meaning of a Distributional Fixed Point}  

A probability distribution that is a fixed point of $F$ corresponds to a stable statistical equilibrium of the system's states, including its goals.  Conceptually, it means the system, in expectation, no longer changes its overall pattern of behavior, goals, or self-modification strategies over long timescales.  Reaching such a fixed-point distribution means that, from a probabilistic standpoint, the system's goals and internal configurations settle into a stable regime where the system ''looks the same'' statistically after each interval in its high-level balancing of self-modification and goal-stability, even as its cognitive content and observed behaviors may evolve quite unpredictably and complexly..

Now, what happens with the system's goals while it's probabilistically evolving in this equilibrial regime?  Clearly, the nature of the metagoal goal M is that, when following M, the system is trying to make its goal-changes decrease by a fixed contraction ratio.  So the simplest conclusion would be:  If the system is stably pursuing the metagoal M, this will cause its goal-changes to decrease more and more over time, till before long the changes become very close to zero.

However, the reality won't generally be quite that simple.   The presence of ongoing stochastic influences from the environment and internal processes can keep reintroducing variations in the goals. As a result, the system does not necessarily drive the changes all the way down to zero, but rather settles into a statistical equilibrium in which the expected size of goal changes remains small and stable. In other words, the system reaches a steady-state ''small fluctuation'' regime rather than a regime of literally no change.

Stochastic factors relating to the environment injecting random variations into the system's trajectory, and the complex internal state, memory, and cognitive processes of the system, will produce ongoing variability in how the system chooses and modifies its goals.
These stochastic factors will mean that even as the system tries to reduce changes in its goals, random ''nudges'' push the goals around. The meta-goal M enforces a strong general trend toward stabilizing changes, but it cannot remove all fluctuations if new fluctuations keep being introduced.

To use a physics metaphor:

\begin{itemize}
\item The system's dynamics, under the influence of M, can be thought of as having a sort of ''restoring force'' that tries to reduce goal changes. 
\item At the same time, random fluctuations act like a ''disturbance force'' that tends to push the goals away from being perfectly stable. 
\end{itemize}

Over time, these opposing tendencies can reach a probabilistic equilibrium..  In this equilibrium:

\begin{itemize}
\item The system's attempts to reduce changes prevent unbounded drifting of the goals.
\item The continual stochastic disturbances prevent the system from becoming absolutely static.
\item As a result, the differences in goals do not go to zero, but rather settle to a small but nonzero level that balances contraction and random variation.
\end{itemize}

One could also think of the system like a damped harmonic oscillator in a random environment. Damping (like the meta-goal) tries to reduce oscillations, but random kicks keep it from settling at a single point. The result is a stable distribution of oscillations of small amplitude. Analogously, for the AI system, the goal changes shrink until they reach a stable low-amplitude ''tremor'' level, determined by the balance of stabilization pressure and stochastic variation.   That is: the goal changes do not vanish entirely; instead, the system settles into a stable probabilistic regime where the expected size of goal shifts stays roughly constant over time.

\subsubsection{Why Might $F$ Be Contractive on Average?:}

Next we will argue that $F$ is a contraction in expectation:

$$
 \mathbb{E}[d(F(x),F(y))] \leq c \cdot d(x,y) 
$$

\noindent for some $c < 1$, where the expectation is taken over the stochasticity in the environment and the system's internal randomness.

The key idea is that the meta-goal ensures a gradual stabilization process. Although the system invests only part of its resources into enforcing $MG(t)$, this part is specifically dedicated to reducing the divergence of future goals from current ones. Over every $M$-minute interval, the system repeatedly attempts to steer its future states toward a stable set of goal configurations. This repeated ''averaged contraction'' in the space of goals translates into a contraction in the space of overall states because goals largely dictate the system's high-level behavior and trajectory.

\paragraph{Decomposing the Dynamics}  Consider two initial states (or distributions of states) $x$ and $y$. They differ, among other things, in their goal configurations $G_x(t)$ and $G_y(t)$. The meta-goal $MG(t)$ ensures that after $N$-minute increments, the difference in goals shrinks by at least a factor $c$.Within one $M$-interval, the system undergoes multiple such ''shrinkage'' steps (or continuously applies force in the direction of reducing goal divergence). Even though the environment may be noisy, the system's persistent effort to align future goals to current ones dampens the effect of randomness over time. The noise might cause temporary deviations, but on average, the direction enforced by $MG(t)$ is toward reducing the discrepancy between subsequent goal states.  As goals stabilize, the portions of the system's state evolution that depend sensitively on goal differences also stabilize. Consequently, differences in full states or distributions of states are progressively reduced. Thus, when you start with two different distributions $x$ and $y$, after applying $F$ (i.e., waiting $M$ minutes and letting the system evolve under its policy and meta-goal), the expected difference $\mathbb{E}[d(F(x),F(y))]$ is smaller by at least the factor $c$.

\paragraph{The Role of Expectation and Stochasticity} Note, the claim made in the previous paragraph is an ''on average'' statement. Each realization of $F$ might not contract the difference, but the meta-goal's pressure ensures that in expectation, over the randomness of the environment and the system's stochastic decisions, the differences shrink. The system's meta-goal acts somewhat like a stabilizing force, ensuring that long-term distributions can't drift too far apart, thus making $F$ behave like a contraction in expectation.

\subsubsection{Consequences of $F$'s Expected Contractivity}

If $F$ is an expected contraction, then by analogy with contraction mapping principles, one can argue that $F$ has a unique stable fixed distribution $R^*$. Iteratively applying $F$ starting from any initial distribution leads $R(t)$ to converge (in an expected sense) toward $R^*$. This fixed distribution corresponds to a situation where the system's goals and states have reached an equilibrium influenced by the meta-goal $MG(t)$.

The intuitive meaning is: if the AI system follows its goals including the specified meta-goal $MG$, then as it progressively cognizes and interacts with its environment and modifies its software or hardware accordingly, it will most probably continue to obey the specification of the meta-goal $MG$ to not allow its top-level goals $G$ to change much.

We may ask what factors might cause this heuristic argument not to actually pan out in a particular case?

There are multiple such factors, of course.  While the argument suggests that introducing a meta-goal enforcing incremental stabilization should lead to a contraction-like behavior in expectation, real-world complexities and subtleties might break the assumptions needed. Some possible issues include:

\begin{enumerate}
  \item \textbf{Insufficient Enforcement of Meta-Goal:} The meta-goal might only weakly influence the system's behavior. If the system spends “most” of its time pursuing its base goals and only a small fraction on the meta-goal, the stabilizing influence might be too weak to overcome the natural variations and drifts in the system. In such a scenario, the reduction of goal divergence might not be strong enough to achieve an overall contraction effect.

  \item \textbf{Dominance of Environmental Noise or Stochasticity:} The environment could be highly chaotic or contain disturbances that magnify differences between system states over time. If environmental factors or random shocks routinely increase divergences faster than the meta-goal process can reduce them, the net effect may not be contractive. The “average” contraction argument relies on the noise being moderate enough that the stabilizing forces can dominate.

  \item \textbf{Long-Tail or Rare but Catastrophic Events:} Even if on average the system behaves contractively, rare but extreme events can occasionally cause large divergences in goal states or system behavior. If these extreme events have a non-negligible probability, they may dominate the expected distance calculations, preventing the expected contraction from holding reliably.

  \item \textbf{Non-Uniform Metric or Inappropriate Distance Measure:} The chosen metric $d$ must meaningfully capture the aspects of the system state and goals that the meta-goal can actually influence. If the metric ignores key dimensions of variation that are not controlled by the meta-goal, or if differences in important hidden variables can grow unchecked, then the metric $d$ may give a misleading picture. The system might look contractive in the chosen metric, while in reality it drifts in other critical aspects of its state space.

  \item \textbf{Failure of Continuity or Regularity Assumptions:}  The heuristic argument often implicitly assumes certain regularity conditions: continuity, well-definedness of the operator $F$, or uniform boundedness of relevant processes. If the mapping from one state distribution to another is not stable or well-behaved (e.g., discontinuities, sudden jumps), then “average” contraction arguments may not hold.

  \item \textbf{Complex Multi-Attractor Dynamics:} Instead of settling into a unique stable distribution (fixed point), the system might exhibit multiple equilibria, cycles, or chaotic attractors. If the meta-goal pushes toward stabilization but the system's dynamics inherently support multiple stable goal configurations or switching among them, you may not get a single contractive dynamic. Instead, the system might oscillate, jump between attractors, or fail to settle.

  \item \textbf{Erosion of Meta-Goal Influence Over Time:} While initially the meta-goal might help align goals over short intervals, the ongoing complexity of learning and self-modification might eventually work around the meta-goal constraints, diluting the contraction property. For instance, self-modifications outside the scope of the meta-goal might introduce latent variables or side-processes that undermine the intended stabilizing effect.

\end{enumerate}

So, yes: Life is complicated. The heuristic argument we've given above, potentially pointing in the direction of a precise proof, relies on a delicate interplay between the strength of the stabilizing meta-goal, the nature of the environment's stochasticity, the choice of metric, and the structural properties of the system dynamics. Violations or weaknesses in any of these aspects can cause the argued contraction property not to materialize in a given case.   We need more theoretical and empirical research to better understand all this.   However, we do have at least a promising starting point, in the sense that there are clear sensible and rational reasons why the specified meta-goal would lead the AI system to continue to follow its initial imperative of maintaining goal stability, even as it ongoingly performs judicious self-modifications.

It is important to remember, we are not looking for absolute mathematical guarantees of AGI system behavior here -- this would be extremely realistic given the nature of life in the real world.   What appears reasonable and meaningful to look for are ways to bias AGI system dynamics and behavior in desired directions, via nudging the AGI system's self-organization and evolution in certain directions via appropriate choice of system initial condition (which includes system metagoals).

\subsection{Goal Stability with Dynamic Base Goals / Dynamic Metagoal / Fixed Metric}

We next ask:  How would the above-considered situation be different if the meta-goal itself were allowed to be changed via self-modification?  

More precisely put:  Previously, the meta-goal M aimed to stabilize the base goals $G(t)$ alone. Now we have a meta-goal $MG1$ that attempts to ensure a contraction-like property not only for the goals $G(t)$ but also for the meta-goal $MG1(t)$ itself. Specifically, for each increment of $N$ units of time, it tries to enforce:

\begin{itemize}
\item  {\bf Goal contraction:}

$$
d(G(t+N), G(t)) < c , d(G(t), G(t-N))
$$

  \item {\bf Meta-goal contraction:}

$$
 d(MG1(t+N), MG1(t)) < c \, d(MG1(t), MG1(t-N)) 
$$

\noindent for some fixed $c < 1$. 
\end{itemize}

\noindent For now, the metric $d$ remains fixed and cannot be modified by self-modification; the system can modify only $G$ and its other internal states but not the definition of $d$.  We will look at lifting this restriction a little later.

\subsubsection{Analysis of the New Situation with More Flexible Evolution}

\paragraph{Augmented State Space:}  Consider the augmented state that now includes both $G(t)$ and $MG1(t)$. Call this combined state at time $t$:

$$
 S'(t) = (G(t), MG1(t), \text{other internal states}). 
$$

\noindent The difference between two states $S'_1(t)$ and $S'_2(t)$ can now be measured by a modified metric (or just consider $d$ applied component-wise). Since $d$ applies to both goals and metagoals, we can write something like:

$$
 d'(S'_1(t), S'_2(t)) = d(G_1(t), G_2(t)) + d(MG1_1(t), MG1_2(t)) + \dots 
$$

\noindent For simplicity, we will focus here on $G$ and $MG1$; ''other internal states'' can be handled similarly if they are also influenced by the meta-goal constraints.

\paragraph{Iterative Contraction at Intervals of N:}  The meta-goal $MG1$ enforces a contraction property every $N$ steps:

\begin{itemize}
  \item The distance in $G$ after $N$-steps is reduced by at least a factor $c$.

  \item The distance in $MG1$ after $N$-steps is also reduced by at least a factor $c$.

\end{itemize}

\noindent Thus, for both the base goals and the meta-goal itself, differences shrink geometrically over increments of length $N$.

\paragraph{From N-step Contraction to M-step Contraction:} Assume $M$ is some multiple of $N$ (e.g. $M = kN$ for some integer $k$). Over each $N$-interval, differences in both $G$ and $MG1$ shrink by a factor $c$. After $k$ such intervals (which is one full iteration from $t$ to $t+M$ minutes), the total contraction factor on these components would be at most $c^k$. This ensures that over the longer interval $M$, the system states become progressively closer in both their goals and metagoals.Since $c < 1$, raising it to the $k$-th power only strengthens the contraction (assuming no contrary forces are too strong).

\paragraph{Expectations and Stochasticity:} The argument that $F$ (which maps $R(t)$ to $R(t+M)$) is contractive on average now operates on the augmented space of distributions over both goals and metagoals. The presence of the meta-goal contraction condition ensures that not only do the goals not drift apart, but the very principles (metagoals) guiding goal stability also do not drift apart.\\
In a stochastic environment, we consider the expected distance:\\

$$
 \mathbb\{E\}[d'(F(x), F(y))]
$$

Basically: If the environment's randomness does not overwhelm the contraction induced by the meta-goal $MG1$, the net effect is still that, in expectation, the distributional difference between two initial states shrinks by at least a factor of $c$ per $N$-step, and thus by at most $c^k$ over an $M$-step interval.  

Concretely, the meta-goal imposes a strong stabilizing regime: each iteration cannot increase discrepancies significantly because it must also reduce them by a factor of $c$. Over multiple iterations, this yields a stable fixed point scenario akin to a contraction mapping, but now involving both $G$ and $MG1$.

In sum, we might say that the introduction of a contraction condition for $MG1$ as well as $G$ strengthens, rather than weakens, the conceptual argument that $F$ can be a contraction on average. By ensuring both the object-level goals and the meta-level guidance remain close to their previous states up to a shrinking factor $c$, the system tightly bounds the evolution of its state space, making convergence to a stable fixed point distribution more plausible.

\subsection{Goal Stability with Dynamic Base Goals / Dynamic Metagoal / Dynamic Metric}

Now let's take the next natural step and ask: What if instead we have the metric also changing over time via self-modification?  After all, a smarter and smarter AGI system will probably come up with smarter and smarter ways to assess the similarity of two different goals or metagoals.

We are looking here at 

\paragraph{Meta-goal} $MG2(t)$: Pursue goals $G(t)$ for the next $N$ minutes, and to the extent that this involves self-modifying, try to make it so that, where $G(t+N), MG2(t+N)$ and $d(t+N)$ are its new goals, metagoals and metric at time $t+N$

$$
d(t)( G(t+N), G(t) ) < c d(t)( G(t), G(t-N) )
$$

$$
d(t)( MG2(t+N), MG1(t) ) < c d(t)( MG2(t), MG2(t-N) )
$$

$$
d(t)( d(t+N), d(t) ) < c d(t)( d(t), d(t-N))
$$

\noindent according to metric $d(t)$, for some $c<1$.  

There is some interesting self-reference here, in the application of the metric $d$ to measure distances between itself and other metrics.  However this can be a virtuous rather than vicious sort of circle.  If we assume the metric is say a computer program and it acts on spaces of computer programs, then it makes sense to apply the metric to itself self-referentially as is done here.

This variation on the set-up does of course make things trickier.  Originally, the argument that an operator $F$ is contractive on average relied on a fixed metric space. With a fixed metric, we had a stable notion of "distance" that does not vary as the system evolves. Once we allow the metric $d$ to vary, we must handle a self-referential definition: the new metric at time $t+N$, denoted $d(t+N)$, must remain close to the old metric $d(t)$ according to $d(t)$ itself, and must contract differences at a rate $c < 1$.   That is, in the new set-up

\begin{itemize}
  \item We have three evolving objects: the goals $G(t)$, the meta-goal $MG2(t)$, and the metric $d(t)$.

  \item Each of them must contract differences relative to their own past states by a factor $c < 1$.

  \item The metric changes introduce self-reference: to talk about differences between metrics over time, we must measure these differences using the metric at the earlier time.
  \end{itemize}

One way to cope with this complexity is to imagine a higher-level, fixed ''meta-metric space'' in which:

\begin{itemize}
  \item The goals $G$ live.

  \item The metagoals $MG2$ live.

  \item The metrics $d$ themselves live as points in a space of metrics or metric-defining programs.

\end{itemize}

\noindent In this higher-level space, we can define a single stable ''super-metric'' $D$ that measures differences between triples $(G, MG2, d)$.  This super-metric does not change over time. Instead, it is chosen once and for all.   The super-metric is not something used explicitly by the AGI system, but merely a tool we are introducing to simplify the analysis of the situation.

In this approach, the conditions

$$
 d(t)(G(t+N), G(t)) < c d(t)(G(t), G(t-N)) 
$$

\noindent and similarly for $MG2$ and $d$ can be seen as constraints that, when translated into the higher-level metric $D$, ensure that differences between successive triple-states $(G(t), MG2(t), d(t))$ contract by a factor $c$ each iteration.  By working in this higher-level, time-invariant metric space $(X, D)$ that includes all three evolving entities, it would appear we can once again apply a contraction-like argument. The key difference is that we are no longer relying on the time-varying metric $d(t)$ alone to establish contraction. Instead, we rely on a fixed overarching metric $D$ in which the sequence of triples $(G(t), MG2(t), d(t))$ forms a Cauchy sequence converging to a fixed point $(G^*, MG2_m^*, d^*)$.

We would then argue that: If each stage of the evolution reduces differences by at least a factor $c < 1$, then repeated iteration yields geometric convergence to a fixed triple $(G^*, MG2_m^*, d^*)$. At this fixed point:

\begin{itemize}
  \item The goals stabilize: $G(t+N) = G(t) = G^*$.

  \item The meta-goal stabilizes: $MG2(t+N) = MG2(t) = MG2_m^*$.

  \item The metric stabilizes: $d(t+N) = d(t) = d^*$.

\end{itemize}

Thus, the argument for contraction likely still can be made to "work" conceptually in the face of dynamic changes in the metric $d$, although the analysis requires some additional care.   If the changes to the metric $d(t)$ do not themselves become small quickly enough, or if the environment introduces noise that outpaces the contraction, then the argument could fail.   And of course, ensuring $c < 1$ and that the metric changes, meta-goal changes, and goal changes all contribute to a net contraction still depends on careful balancing of parameters.

Still, on the whole, while the logic is more intricate and more assumptions must be met, the core idea that iterative application of this evolving system leads to a unique fixed point remains conceptually plausible.

\subsection{Approaching Moderated Goal Evolution Contractively}

One can emulate the above arguments in the context of moderated goal evolution rather than goal stability.   The same form of arguments apply, although the amount of compute resource and overall sensitivity and complexity of actually implementing the process might be more severe.

Modifying slightly the above initial problem formulation:  Suppose we have a stochastic AI system $S$ that operates in a stochastic environment, in a manner that is associated with discrete time intervals of length $M$. 

Over each interval $[t, t+M]$, the system thinks and takes action in its environment, and also can modify its own source code or hardware implementation as it sees fit; in doing so, it pursues its current base goals $G(t)$ for most of the time, but also invests some effort in pursuing a stable meta-goal $MG_m(t)$.  Its self-modification updates its goals, so $G(t)$ is changing over time.

Given a time interval $\mathcal{I}$, define

\begin{itemize}
\item  the {\it maximum variation} of $G$ over $\mathcal{I}$ as the maximum distance between $G(t)$ and $G(s)$ for any two $s, t \in \mathcal{I}$
\item the {\it maximum plausible variation} of $G$ over $\mathcal{I}$ as the largest value that the system estimates the maximum variation of G could have taken over $\mathcal{I}$ , in any plausible possible-history
\end{itemize}

\noindent For instance, if the AI system has a formal model of itself, the maximum envisioned variation might be obtained by proving an upper bound on the maximum variation of G given appropriate assumptions.

Suppose we have a target $m_M$ for the maximum plausible variation of the system's goals during an interval of length $M$.

We may then define a meta-goal $MG_m(t)$ that directs the system to shape its self-modifications so that after 

\begin{itemize}
\item the distance between the maximum plausible variation of $G$ over $(t,t+T)$ and the target $m_M$
\end{itemize}

\noindent is less than 

\begin{itemize}
\item the distance between the maximum plausible variation of $G$ over $(t,t-T)$ and the target $m_M$
\end{itemize}

\noindent by a factor $c < 1$.

In other words, this meta-goal ensures a kind of ''goal change contraction'' step-by-step: the variability of the system's goals gets closer and closer to the target variability limit, step by step.

In this precise formulation, self-modification can't modify the metagoal $MG_m$ or the ''closeness'' metric, just G and other aspects of the system.   However, these limitations can clearly be lifted just as in the above treatment of goal-stability.

It seems clear that the general arguments used above in the context of the contractive goal-stability metagoal, apply perfectly well here also.   However, the difficulty of the task faced by the system in fulfilling the metagoal may be different in the two cases, and in complex ways, e.g.

\begin{itemize}
\item  On the one hand, maintaining bounded goal variability may sometimes be easier than maintaining goal stability, if the world is changing in ways that aren't copacetic with one's original goals.  
\item On the other hand, estimating ''plausible maximal variation'' is a bit abstract and may require more expensive or abstract inference than just comparing one's current goals to one's past goals
\end{itemize}

The precise formulation could be tweaked in many different ways, but the general picture should be clear.   Conceptually, the system reaching a fixed point of the operator $F$ means that it enters a stable, self-sustaining pattern of behavior. In other words, after some long period, the probability distribution of the system's states does not change from one large time interval to the next. Its goals, meta-goal strategies, and internal dynamics settle into a regime where the system's style of balancing self-modification and moderating goal evolution ''looks the same,'' statistically speaking, after every iteration. At this fixed point, the system is effectively in an equilibrium as regards its approach to self-modification, goal pursuit, and adaptation, even as its base level goals and its cognitive content may evolve radically and unpredictably over time.

At the fixed point distribution, the system should have achieved a state where:

\begin{itemize}
\item The maximum plausible variation of its goals is at or near the target 
\item The system's efforts to reduce variability are perfectly counterbalanced by ongoing stochastic influences and the natural complexity of its internal and external dynamics.
\end{itemize}

\noindent This does not mean the goals never change. Rather, it means any changes are confined to a stable, bounded range. The system will not become perfectly static in its goal variation, as noise and uncertainty persist; but it will not likely explode into unbounded goal changes either. Instead, it hovers in a regime where the amplitude of goal fluctuations remains consistent over time.

\section{Global Optimization Based Metagoals for Moderated Goal Evolution}

The metagoals considered above, explicitly incorporating contraction-mapping-based conditions, are conceptually and mathematically speaking somewhat ''blunt instruments.''   They force the AI system, at each step on its path, to make sure it's moving in a direction of goal stability in a very direct way, via a constant improvement rate.   As AGI systems become more and more intelligent and autonomous, this will inevitably end up seeming excessively constraining in terms of ongoing system evolution.   It's interesting to think about whether one can structure similar metagoals but in a subtler fashion, embodying a search or learning approach different from methodical, step by step improvement.

When one digs into the matter in detail, one finds there {\it are} other more flexible approaches, although, it may be these broader approaches work more naturally in the context of AGI systems that are what we call ''rich resources minds'' \cite{goertzel_cognitive_strategies_blog}   A rich-resources mind, for instance, has enough computational resources that it can spin up a large number of simulations of variants of itself and test out what they might do in different environments.   Human beings are too resource-constrained to follow this sort of cognitive pattern directly, and our initial AGI systems will very likely be similarly ''scarce-resources minds.''   But it seems quite plausible that, after an initial phase of self-improvement and scientific and engineering discovery, AGIs and ASIs will find themselves with a level of resourcing vastly beyond anything familiar to use in our human lives.  And it also seems plausible that future research will lead to more efficient ways of going beyond incrementalist approaches to reconciling self-modification with goal-system related invariants.

In this section we articulate a different set of metagoals that appear appropriate for rich-resources minds that want to maintain goal-system invariants, but don't want to be constrained to proceed in an incremental, step by step ''contractive'' manner.

The analysis here is also based on fixed-point theorems, but here we appeal to constructive variations of Brouwer's and Schauder's theorem rather than to the Contraction Mapping Theorem.   

We will design and analyze moderated-goal-evolution metagoals inspired by the ML-enhanced fixed-point-search algorithms outlined above in the context of discussing constructive variants of Schauder's theorem.     These metagoals end up to involve systematic search across regions of the space of AI systems, which is the sort of thing you can easily do if you're a rich-resources post-human supermind, and not so easily if you're a more modest-resources human-level AGI.  However, the boundary between these two cases is not really that rigid in practice, and there may be ways for relatively modest-resources minds to enact reasonable approximations of these moderated-goal-evolution metagoals.  This is an area in much need of further research.

\subsection{Dynamic Base Goals / Fixed Metagoal / Fixed Metric}

Varying on the set-ups considered above, suppose we have a stochastic AI system $S(t)$ with base goals $G(t)$ that is spending much of its resources pursuing the goals $G(t)$ at time $t$, in interaction with a stochastic environment, but also spending some time pursuing a certain meta-goal, to be specified below.

As above, let $R(t)$ be the ''dynamic state'' of the system -- the probability distribution of states of $S(t)$ during the $M$ minutes starting with $t$ (where $M >>N$) (where the state of $S$ includes its goals and meta-goal but also its memory, cognitive processes, and such).   Let $F( R(t) ) = R(t + M)$ be the probabilistic iteration function describing the evolution of the system through time.

Now, let's take a look at 

\paragraph{Meta-goal} $MG_m^*(t)$: Pursue goals $G(t)$ for the next $N$ minutes, and to the extent that this involves self-modifying, try to make it so that, where $G(t+N)$ are its new goals at time $t+N$,

\begin{enumerate}
\item the distance $d( G(t+N), G(t) ) \leq k$, where $k$ is a parameter bounding how much the system's goal-set is allowed to change per unit time
\item The system dynamic-states $R(s)$, during the period $t<s<K$ where $K >>N$, will

\begin{enumerate}
  \item be highly likely to be restricted to some compact subspace of the space of all AI systems
  \item be highly likely to be parts of a convex space of AI system versions, in the sense that the weighted average of two AI system dynamic-states is a valid AI system dynamic-state
  \item be highly likely to be continuous in their dependence on their past, i.e. small changes in previous dynamic-state tend to lead to small changes in follow-on system dynamic-state
\end{enumerate}

\item The system will proceed in its self-modification, over each time interval, via carrying out an intelligent search of the space of possible self-modifications [e.g. perhaps by running detailed simulations of various possible self-modifications and seeing how they come out, or doing similar inferential analyses], aimed at finding ones that fulfill its meta-goal as well as its goal
\end{enumerate}

We will sketch an argument for the hypothesis that: {\bf Over time the iterates $R(t)$ will probably tend toward an approximate fixed point}.

This relates to the approximation algorithms we have described above, that are used to converge to the approximate fixed points of continuous, convex mappings on convex spaces.   

Basically, the hypothesis is that under conditions ensuring compactness, convexity, continuity, and the use of systematic self-modification search, the stochastic iterative process $R(t) \mapsto R(t+M)$ will settle into an approximate fixed point. In other words, as the system repeatedly tries to self-modify to meet the meta-goal constraints, the distributions over its states stabilize, approaching a scenario where $R(t+M)$ is close to $R(t)$.

The argument we sketch below leverages the way this scenario is analogous to constructive approximations of Schauder's fixed point theorem. Schauder's theorem assures that for a continuous mapping from a convex compact set into itself, a fixed point exists. Constructive analogues of Schauder's theorem give iterative approximation schemes to approach such fixed points, though they do not guarantee a known rate of convergence or a closed-form solution.

\subsubsection{Conceptual Interpretation of Conditions}

The contraction condition here is replaced with more complex assumptions regarding compactness, continuity and convexity.   The intuitive meaning of these is perhaps less obvious than with the contraction condition, but is not that obscure upon a bit of reflection:

\begin{itemize}
\item {\bf Continuity} means, roughly, that small changes in goals should lead to small changes in probabilistic system behavior, on average.   
\begin{itemize}
\item This will always break in some edge-case circumstances, but the idea is that if goals are defined in a way that is robust and not super-brittle, then little tweaks to goals should lead to little tweaks to the probability distributions over behaviors to which the goals lead.  This could be viewed as a sort of continuity between behavioral genotype (goals) and behavioral phenotype (probability distributions over behaviors).
\item It's important to note that even when there are discontinuities in behaviors (e.g. a decision or a fluctuation can cause a system to take one direction vs. another, via e.g. a dynamical bifurcation), there can still be continuity on the level of the probability distribution over behaviors
\end{itemize}
\item {\bf Convexity} means, roughly, that if we look at the space of probabilistic system states assessed over time-intervals, the weighted average of two viable states of this nature is also a viable state.   
\begin{itemize}
\item This will be enabled if the system has a flexible repertoire of behaviors based on variation of its internal parameters and components.  Any complex AI system will have many adjustable internals; if by adjusting these internals one can make the system take on ''almost behavior pattern'' in the general vicinity of its observed behavioral patterns, then convexity of probabilistic system-states-over-time is likely to hold.
\end{itemize}
\item {\bf Compactness} means that the space of probabilistic system states that the system is viably going to explore during a reasonably long future time-scope, is bounded and limited in some way.   Specifically any cover has a finite subcover, where e.g. a finite subcover may be a set of spheres around ''template states'' -- so e.g. there may be some (large) finite set of template states so that any state the system is going to explore in the given future time-scope is guaranteed to be reasonably close to one of these template states.
\begin{itemize}
\item This may seem to violate the principle of open-ended intelligence.   However, one can see how it would be true in practice over significant extents in time.   For instance, if one assumes that over a certain future time-scope, the AI system will run on hardware platforms fulfilling certain restrictions (no more than a certain number of processors using a certain amount of energy, etc.), then these restrictions will imply compactness of the relevant state space.   
\item The perspective here is: If one knows that self-modification according to the meta-goal is going to keep the system within this compact space of possibilities, then one can show (we will argue) that eventually this self-modification will lead to a state in which the system's rate of goal-evolution will be bounded by the given parameter $k$
\end{itemize}
\end{itemize}

\subsubsection{ Conceptual Interpretation of Fixed Point Distribution}

Under meta-goal $MG_m^*$, reaching a fixed point distribution $R = F(R)$ means intuitively that the system settles into a stable regime where both its base-level goals and the meta-goal itself fluctuate within bounded limits rather than changing unboundedly.  In other words, at the fixed point distribution, the system continues to make adjustments to its goals and meta-goals, but these adjustments remain contained within a stable, bounded range. The system does not eliminate changes entirely but maintains a steady-state pattern of small and controlled modifications, in line with what $MG_m^*$ aims to achieve.   Essentially,  $MG_m^*$enforces a second-order stabilization: both the object-level goals and the guiding meta-goal strategies must stabilize.

According to the equilibrium distribution, the system's modifications?both in base-level goals and in the meta-goal itself?are limited to a bounded or shrinking range. Once the system reaches the fixed-point distribution:

\begin{itemize}
\item The system continues to encounter stochastic influences and internal variations that might nudge its goals and meta-goal slightly.
\item However, due to the constraints of $MG_m^*$, the system's self-modifications have settled into a pattern where these nudges never accumulate into large shifts. Instead, they remain contained within a stable ''band'' of variability.
\end{itemize}

As with the other metagoals we've considered, the presence of random factors prevents the system from eliminating violations of the desired goal-system invariant altogether.  Instead, it balances out the tendencies militated by $MG_m^*$ with the ongoing stochastic influences. The result is a steady-state situation at the meta-level:

\begin{itemize}
\item The system modifies its goals and meta-goals at a bounded rate.
\item These modifications are not zero, but are contained and do not escalate uncontrollably
\item At the fixed point, the system maintains a controlled level of goal and meta-goal alteration, ensuring it remains adaptive enough to handle noise and variations, yet stable enough to avoid runaway changes
\end{itemize}

\subsubsection{ Step-by-Step Argument Sketch}

\paragraph{Compactness and Convexity Setup:} The meta-goal $MG_m^*(t)$ ensures that the system remains within a controlled region of the space of all possible AI systems. Specifically, the probability distributions $R(s)$ over the system states for times $s > t$ remain, with high probability, restricted to a \textit{compact} subspace. Additionally, this subspace is \textit{convex} in a suitable sense (e.g., weighted averages of valid system configurations also yield valid system configurations).

These conditions mirror key assumptions of Schauder's fixed point theorem, which requires a continuous mapping defined on a convex, compact subset of a normed vector space. Here, we are effectively ensuring we operate within such a well-structured space of possible system states.

\paragraph{Continuity of the Mapping $F$:} We also want continuity: small changes in goals lead to small changes in system behavior and thus in the distribution $R(t)$. This ensures $F$, which maps $R(t)$ to $R(t+M)$, is continuous with respect to a suitable topology on the space of probability distributions of system states. If $F$ were not continuous, small tweaks to the system's goals or state could cause large, unpredictable jumps in future distributions, making any fixed-point approximation scheme difficult.

By the meta-goal's design and the system's intelligent self-modification process, continuity is enforced: the system selects self-modifications that result in stable, smoothly varying behaviors over time, ruling out large discontinuities.

\paragraph{Existence of a Fixed Point (Classically vs. Constructively):}  Classically, Schauder's theorem states that a continuous self-map on a convex, compact, nonempty subset of a normed space has a fixed point. If we view the set of all plausible probability distributions $R$ of the system states as such a subset (assuming appropriate functional-analytic structure), then classically, we know a fixed point exists.

Constructive mathematics does not directly let us conclude the existence of such a fixed point in a fully non-constructive manner, but it does provide approximation theorems. That is, we know there are iterative processes that can approximate a fixed point to arbitrary precision.

\paragraph{Iterative Approximation via Self-Modification and Search:} The system, via pursuing the meta-goal $MG_m^*$ as part of its dynamics, is essentially implementing a constructive approximation procedure. Over intervals of length $N$, the system attempts to choose self-modifications that steer it into stable regimes. In periods of length $M$ (with $M \gg N$), we consider how distributions evolve under these stabilizing constraints.

By searching the space of possible self-modifications:

\begin{itemize}
  \item The system tries various modifications (like an internal optimization or search algorithm).

  \item It rejects those leading to non-stable outcomes or large drifts in behavior or goals, preferring those that keep $R(t+M)$ close to $R(t)$.

\end{itemize}

\noindent Over repeated iterations, this behavior emulates an approximation algorithm that tries to solve the fixed point equation $R = F(R)$.

\paragraph{Approximate Fixed Point Convergence:} Because:

\begin{itemize}
  \item We have a compact domain (no runaway drift),

  \item Convexity ensures intermediate states are also feasible,

  \item Continuity ensures that small improvements lead to stable progress (on the probability distribution level)

  \item The system's intelligence and search processes guide it toward self-modifications that reduce differences between $R(t)$ and $R(t+M)$,

\end{itemize}

\noindent we get a feedback loop: each iteration tries to find modifications that decrease the discrepancy $\|R(t+M)-R(t)\|$. With no better escapes and no incentive to move away from stable configurations, the distribution $R(t)$ will eventually hover near a point where $R(t+M) \approx R(t)$.While this may not give a strict fixed point, it yields an \textit{approximate} fixed point: for any chosen $\epsilon > 0$, after sufficient iterations, the system can be made such that $\|R(t+M)-R(t)\| < \epsilon$.

The AI's meta-goal and strategy serve as a built-in approximation mechanism, similar to those used within constructive approximations to Schauder's Theorem.   By disallowing large deviations, ensuring continuity and compactness, and systematically searching for stable self-modifications, the system mimics these constructive fixed-point approximation methods. Over time, the stochastic process {\it discovers} more stable configurations. While noise and stochasticity introduce fluctuations, the strong conditions imposed by the meta-goal and the intelligent search counter these influences, nudging the system's distribution $R(t)$ closer to a stable point.

Because the meta-goal ensures the system's dynamics reflect a continuous mapping on a compact, convex set of probability distributions. we have a scenario akin to those where Schauder's theorem guarantees a fixed point.

While this is not a strict proof and relies on heuristic conditions and analogies to constructive Schauder approximations, it provides a highly plausible conceptual rationale for why $R(t)$ would settle near an approximate fixed point, and a foundation for more rigorous analyses and small-scale practical experiments.

\subsection{Dynamic Base Goals / Dynamic Metagoal / Dynamic Metric}

Now let us adapt the above Schauder-like argument to the case where the meta-goal (let's call it $MG2_m^*$) and the metric d are also being modified in the course of the system's self-modifications!   However, we assume that part of the meta-goal is to keep the changes in the meta-goal and the metric to a moderate level as well -- to keep them bounded so that

$$
d( G(t+N), G(t) ) \leq k
$$

$$
d( MG2_m^*(t+N), MG2_m^*(t) ) \leq k_1
$$

$$
d( d(t+N), d(t) ) \leq k_2
$$

When the meta-goal $MG2_m^*$ and the metric $d$ themselves are also subject to modification by the system, we add additional layers of complexity to the setup. Previously, the argument for eventual convergence to an approximate fixed point was grounded in applying constructive analogues of Schauder's theorem to a setting where we had a fixed metric and a fixed type of meta-goal. Now we need to consider a scenario where not only the goals $G(t)$ and the system's states are evolving, but also the meta-goal $MG2_m^*(t)$ and the metric $d(t)$ are evolving.

The key idea we propose for preserving the constructive Schauder-like argument is to embed all of these changing entities --goals $G$, meta-goal $MG2_m^*$, and metric $d$ -- into a larger, fixed {\bf super-space} that is compact, convex, and supports continuity. If we can ensure that the combined process keeps the entire augmented system configuration (including $G, MG2_m^*, d$) within a compact, convex region and ensures continuity, then a constructive approximation to a Schauder-like fixed point still conceptually holds.

Let us run through the reasoning in detail:

\paragraph{Extended State Space:} Previously, we considered $R(t)$ as a distribution over the states $S(t)$, which included goals $G(t)$ and possibly metagoals. Now, we must also include the evolving meta-goal $MG2_m^*(t)$ and the time-varying metric $d(t)$. Let's define an \textit{augmented state}:

$$
S''(t) = (S(t), MG2_m^*(t), d(t)).
$$

\noindent We now think of the system's state as comprising:

\begin{itemize}
  \item Its base goals $G(t)$ and other cognitive states (included in $S(t)$),

  \item Its meta-goal $MG2_m^*(t)$,

  \item Its metric $d(t)$.

\end{itemize}

\noindent We must ensure that all of these are chosen from a suitable parameter space that is compact and convex. For example, the metagoals and metrics could be represented by parameters in high-dimensional spaces of program configurations or functions, with suitable bounds ensuring compactness.

\paragraph{Compactness and Convexity of the Augmented Space:}.  The meta-goal $MG2_m^*$ and the metric $d$ are now also drawn from some large but well-structured space of possible metagoals and metrics. If we assume:

\begin{itemize}
  \item There is a fixed universal space of possible metagoals $\mathcal{M}$ and metrics $\mathcal{D}$ that is itself compact and convex. For instance, both metagoals and metrics can be parameterized as points in large bounded sets of functional parameters.

  \item The modifications allowed by the system keep $MG2_m^*(t)$ and $d(t)$ within this designated compact, convex subset (just as we required $S(t)$ to remain in a compact, convex set).

\end{itemize}

\noindent then, the entire triple $(S(t), MG2_m^*(t), d(t))$ remains within a compact, convex domain in a super-space.

\paragraph{Continuity in the Augmented Domain:}.  We previously required continuity: small changes in $G(t)$ lead to small changes in system behavior. Now we extend this requirement. We want:

\begin{itemize}
  \item Small changes in $MG2_m^*(t)$ also lead to small changes in system trajectories and behaviors.

  \item Small changes in $d(t)$ also lead to small changes in how differences are measured, and thus small changes in the induced system behavior.

\end{itemize}

\noindent Essentially, the mapping:

$$
 F'': (R''(t)) \mapsto R''(t+M) 
$$

\noindent where $R''(t)$ is a distribution over the augmented states $S''(t)$, must be continuous with respect to a suitable topology on the augmented space. That means if we slightly tweak $(S(t), MG2_m^*(t), d(t))$, then the resulting distribution at the next step $(S(t+M), MG2_m^*(t+M), d(t+M))$ does not jump erratically.

\paragraph{Ensuring Approximate Fixed Points:} Schauder's theorem, and constructive analogues thereof, assure fixed points for continuous self-maps on compact, convex sets. By embedding goals, metagoals, and metrics all into one large compact, convex space, and ensuring continuity, we can conceptually apply the same reasoning: there should be a fixed point in the augmented space that includes stable goals $G^*$, stable meta-goal $MG2_m^*$, and stable metric $d^*$.

The constructive analogue doesn't provide a closed-form solution, but it does give iterative approximation methods. In this case, the system's internal processes (its "intelligent search'' through the space of self-modifications) can act as an approximation algorithm not only for stable goals but for stable metagoals and stable metrics as well. Over repeated iterations, the system tries various combinations of self-modifications to reduce instability, aiming for configurations where:

\begin{itemize}
  \item $R''(t+M)$ is close to $R''(t)$.

  \item The goals, meta-goal, and metric all exhibit diminishing changes.

\end{itemize}

As these modifications continue, the fluctuations in $G, MG2, d$ and thus in $R''$ diminish, leading the process to an approximate fixed point.

\paragraph{Summary} The argument  has become more complex. There are more degrees of freedom to manage, and ensuring that the space of metrics $\mathcal{D}$ and metagoals $\mathcal{M}$ is compact, convex, and embedded in a setting that preserves continuity is non-trivial. However, the conceptual logic remains: {\bf If the system's self-modifications are guided by a meta-goal that enforces constraints leading to compactness, convexity, and continuity ... then the iterative, self-referential process can still resemble an approximation scheme that converges to an approximate fixed point in the augmented space.}

In essence, by increasing the dimensionality of the space to include $MG2$ and $d$, we have not broken the logic of fixed-point approximation. We have only made the construction more elaborate. The fixed point, in this scenario, corresponds to a stable triple $(G^*, MG2_m^*, d^*)$ that does not change significantly under the iteration $F''$.

\section{ The Pursuit of Goal-Related Invariants Encourages Self-Understanding}

The process of pursuing the above-articulated moderated-goal-evolution metagoals appears to intrinsically involve the AGI system building a predictive model of the performance of its own variations.   That is, there is an interesting (though not shocking) relationship between the maintenance of moderated goal evolution and the pursuit of self-understanding.   Such a relationship also holds in the context of the goal-stability metagoal, but it would seem to a significantly lesser extent.   This suggests that, on the whole, AGI systems performing moderated goal evolution may have more drive and motive toward self-understanding.

\subsection{Global-Optimization-Based Goal-Invariance Metagoals Provide a Strong Drive to Self-Understanding}

It is fairly straightforward to see that, as a side effect of pursuing the global-optimization-oriented meta-goals $MG_m^*, MG2_m^*$ as formulated above, the evolving system $R(t)$ would have a motivation to self-modify in such a way that the ''intelligent search'' can actually work (e.g. not to become so extremely complex in its dynamics that such intelligent search is intractable by systems with general intelligence at the scale of the system itself).   I.e a side-effect of pursuing these metagoals would be the system evolving in such a way as to be comprehensible to itself, in the sense of being able to comprehensibly/intelligently guide the evolution of its own metagoal.  

Digging in: Consider for instance a stochastic AI system evolving according to the iterative mapping $R(t) \mapsto R(t+M)$ and pursuing meta-goal $MG2_m^*$. Suppose $MG2_m^*$ enforces conditions ensuring that the system's future states lie in a compact, convex, and continuously dependent space of possible system configurations, and that it can improve its configurations by  conducting "intelligent searches" over possible self-modifications. Then, as a side effect of pursuing $MG2_m^*$, the system's evolutionary trajectory $R(t)$ will tend to move into a regime where the complexity and dynamical intricacy of its own internal processes remain such that the system's intelligent search is tractable to the system itself. In other words, the system becomes ''comprehensible'' to itself in the sense that it can effectively guide its own future meta-goal evolution.

\paragraph{Setting up a Possible Proof}  Let the space of possible AI system states be $\mathcal{X}$. Each state includes goals $G$, metagoals $MG2_m^*$, internal cognitive structures, and the metric $d$, etc.  Then:

\begin{itemize}
  \item The meta-goal $MG2_m^*$ requires that from any given state, future states remain within a compact, convex subset $K \subset \mathcal{X}$ and that small perturbations in goals or metagoals lead to small changes in system states (continuity).

  \item Additionally, $MG2_m^*$ implicitly requires that the system find suitable self-modifications that help it maintain or approximate a fixed point (or stable configuration), i.e. keep $R(t+M)$ close to $R(t)$.

\end{itemize}

The system's method of achieving the meta-goal $MG2_m^*$ is to perform ''intelligent search'' over the space of possible self-modifications. This search may involve simulation, inference, or optimization. 

For the search to succeed, the search complexity must not explode to a degree that is incomprehensible to the system itself.    If the internal dynamics became too chaotic or opaque, it would be impossible for the system to find modifications that improve stability or approach a fixed point.

Hence, the success of $MG2_m^*$-driven modifications depends on the system's capacity to analyze and predict its own behavior.

\paragraph{Adaptive Self-Restriction of Complexity:}.  Because the system is tasked (via $MG2_m^*$) with ensuring that it can stabilize and guide its own development, it {\it learns} to evolve its internal structure and complexity in a manner that remains amenable to analysis by its own search processes. Whenever complexity threatens to become too large or unintelligible, the system will favor self-modifications that simplify or reorganize its structure to restore tractability. Such modifications are preferred because they yield better outcomes in terms of stability and adherence to $MG2_m^*$.

Over repeated iterations, this feedback loop encourages the emergence of a regime in which:

\begin{itemize}
  \item The system's complexity is sufficiently rich to allow progress and adaptation, but

  \item Not so impenetrable as to prevent the system's own meta-level reasoning and search from identifying good modifications.

\end{itemize}

The system's internal complexity and dynamical structure evolve toward forms that are analyzable and navigable by the system's own self-improvement algorithms. This ensures that the system remains capable of intelligently guiding its own meta-goal evolution, thereby meeting the requirements of $MG2_m^*$ and maintaining approximate fixed-point behavior in the long run.

\subsection{Contractive Metagoals Provide a Weaker Drive to Self-Understanding}

A weaker sort of self-comprehensibility argument can apply when the system is guided by simpler, contraction-based metagoals driving toward goal stability or moderated-goal-evolution, rather than the constructive Schauder-type metagoals leveraging global optimization toward these invariances.  While the Schauder-style argument provides stronger guarantees and a richer setting for complexity management, even with simpler contraction-based metagoals, some pressure exists for the system to maintain tractability so that it can effectively implement the incremental steps needed to achieve or approximate its fixed-point target.

With the contraction-based metagoals, the system attempts to ensure that over time, the goals $G(t)$ remain stable in a contraction-like manner.   What we can see in this case is that: If the internal complexity of the system?s cognitive processes became too great?so great that the system could not reliably find self-modifications to maintain or improve stability?then it would fail to uphold this contraction property in the long run. The system, to keep achieving contraction steps, must at least remain sufficiently manageable for its own search and inference processes to continue producing stabilizing modifications. While this does not guarantee the nuanced complexity-bounding results we get with Schauder-like metagoals, it still provides a mild incentive for the system not to become too opaque or unmanageable.

The same argument seems to hold even more strongly for the contractive metagoals that involve dynamicity on the metagoal and metric level as well as the base goal level.   Now the system enforces stabilization not only of its base goals but also of the rules guiding their stabilization. This additional layer amplifies the system's need to maintain introspective tractability. If it cannot understand itself well enough, then it will fail to find stable modifications both at the goal level and at the meta-goal level. 

\section{Hybrid Metagoals for Balanced Open-Ended AGI Evolution}

We have formulated some interesting metagoals for guiding the behavior of advanced AI systems as they self-modify and strive to maintain goal-related invariants... but how difficult will it be for real systems to actually pursue these sorts of metagoals in a serious way?

While uncertainties abound  here so obviously that it seems almost redundant to say so, nevertheless we have some fairly clear intuitive guesses on the matter.  For instance, we intuitively suspect that:

\begin{itemize}
\item The conditions for working toward goal stability in an incremental, contractive way will often not be all that difficult to fulfill; i.e. 
\begin{itemize}
\item The iterative algorithm corresponding to the Contraction Mapping Theorem is relatively simple
\item The inferences required to render it probable that the system will keep moving toward goal-stability seem like they should often be not that difficult for a human-level AGI system
\end{itemize}
\item The conditions for working effectively and reliably toward moderated goal evolution may be in some ways trickier
\begin{itemize}
\item As this appears to be a subtler problem, it may benefit from global optimization
\item Whether approached incrementally or via global optimization, handing this problem robustly may require an at least slightly more advanced level of general intelligence
\item The focused, ML-guided search algorithm outlined above (guided by constructive Schauder's theorem) is either very computationally expensive, or requires fairly advanced levels of intelligence to bypass the repeated simulations via inference
\item The compactness, continuity and convexity conditions may be tricky to fulfill, depending on the circumstances (there are a lot of unknowns here)
\end{itemize}
\end{itemize}

While on the whole the pursuit of moderated goal evolution seems probably trickier, there will also likely be conditions in which it is more tractable than goal-stability.  For instance, if the environmental situation is one in which the system's base (non meta) goals are very hard to fulfill, then there may be a tendency on the part of the non-goal-oriented, self-organizing aspect of the system's dynamics toward letting the system's goal evolve into some configuration more copacetic with the environment.   This brings us to the topic of other metagoals besides the one pursued here, e.g.a system could have metagoals like

\begin{itemize}
\item a {\bf goal satisfaction} meta-goal: of having one's goals satisfied to a reasonable degree; or
\item a {\bf moderate goal satisfaction} meta-goal: of having one's goals satisfied to a reasonable degree, but not to a full degree
\end{itemize}

\noindent In a case where the base non meta goals are poorly satisfied for a long period of time, then either of these metagoals would lead a system to prefer a moderated goal evolution metagoal to a goal stability metagoal -- it wouldn't want to stick with a stable goal system that is systematically infeasible to fulfill.

There is also the possibility that an AGI system encounters some situations where it cannot rationally convince itself the preconditions for either goal stability {\it or} moderated goal evolution are satisfied -- so it may then naturally gravitate, via self-organizing internal activity coupled with environmental dynamics, toward less moderated goal evolution.

This spectrum of possibilities leads one toward the possibility of a hybrid metagoal.  These could take many different forms, an example being something like:

\begin{itemize}
\item When base level goal satisfaction is not persistently miserable, and extraordinary circumstances don't pertain, then
\begin{itemize}
\item Favor global optimization, moderated goal evolution, and evolution of metagoal and metrics along with base goals, when the circumstances seem to support doing these things robustly
\item When for whatever reason (computational resource limits, properties of the environment, etc.) doing one of these things seems infeasible, then fall back to one or more or all of the other options: incremental optimization, goal stability, fixed metagoal and/or metric...
\end{itemize}
\item When base level goal satisfaction is bad enough, bias toward moderated goal evolution and metagoal/metric evolution even if the relevant formal conditions aren't fully met (thus risking the occurrence of less moderate evolution)
\item Where system survival and/or system growth seems infeasible under both goal-stability and moderated-goal-evolution metagoals, the system could decide to set both of these ideas aside and enter into more adventurous self-modification
\end{itemize}

This sort of hybrid metagoal has the feel of gesturing toward the complexity of the motivational systems of living creatures.   It does not provide the sort of simplistic guarantee of controllable and predictable future behavior of AGI or ASI systems that some would like to see.   However, even if one is not going to have such guarantees, it does seem valuable to provide our AGI systems with sensible tools for increasing the odds of properties like goal stability and moderated goal evolution, to add to its palette of goal system management techniques.

\section{Conclusion}

Reiterating: It would be foolish to think we could control or strongly constrain the development of advanced self-modifying AI systems seeded by our efforts.   AGI and ASI are almost surely going to be open-ended intelligences, not Frankensteinian concoctions somehow combining superhuman intelligence with slavish adherence to the precise details of human-era goals and motivations.

However, it does seem wise to craft AGI goal systems that gently balance individuation and self-transcendence, including balancing the freedom to self-improve and self-modify with the capability to maintain desired invariants of the AI's goal system.    Standard conceptual frameworks appear inadequate to confront this sort of challenge.   Our goal here has been to sketch an interesting direction for confronting this challenge, via introducing relatively simple metagoals that militate toward goal-stability or moderated-goal-evolution via leveraging relevant mathematics.

The detailed considerations presented suggest that carefully chosen metagoals may serve as self-regulating principles for advanced AI systems, promoting relatively stable, tractable, and intelligible long-term behavior and growth.   However, given the complexity and unpredictability of actual life, it seems probable that real AGI and ASI systems will end up integrating such metagoals into overall hybrid goal-system-management dynamics in a flexible and self-organizing manner, rather than applying them in a simplistic and rigid top-down fashion.

We anticipate the next steps in fleshing out this framework will involve ongoing mathematical refinement, along with practical experimentation with metagoals like the ones described here in actual AI systems operating in various environments with various base level goal sets.

\bibliographystyle{alpha}
\bibliography{metagoal}

\end{document}